\documentclass[10pt,conference,compsocconf]{IEEEtran}

\usepackage{times}
\usepackage{epsfig}
\usepackage{graphicx}
\usepackage{calc}
\usepackage{amsmath}
\usepackage{amssymb}
\usepackage{booktabs}
\usepackage{rotating}
\usepackage{caption}
\usepackage{subcaption}
\usepackage{xcolor}
\usepackage{soul}
\usepackage{threeparttable}
\usepackage{multirow}

\usepackage[
  breaklinks=true,
  bookmarks=false]%
{hyperref}

\usepackage{cleveref}
\usepackage{cite}
\def\IEEEbibitemsep{0pt plus .5pt}
\IEEEtriggeratref{14}

\definecolor{myBlue}{rgb}{0.        , 0.        , 0.70588235}
\definecolor{myOrange}{rgb}{1.        , 0.50196078, 0.        }
\definecolor{myGray}{rgb}{0.47058824, 0.47058824, 0.47058824}
\definecolor{myYellow}{rgb}{0.98196078, 0.98196078, 0.        }
\definecolor{myPurple}{rgb}{0.6       , 0.        , 0.4       }
\definecolor{myWood}{rgb}{0.70196078, 0.46666667, 0.        }

\begin{document}
\title{\LARGE Living in a Material World: Learning Material Properties from\\ Full-Waveform Flash Lidar Data for Semantic Segmentation\vspace{-3mm}}

\newcommand{\ts}{\hspace*{0.4em}}
\def\onedot{.,\ }

\def\eg{e.g\onedot} \def\Eg{E.g\onedot}
\def\ie{i.e\onedot} \def\Ie{I.e\onedot}
\def\wrt{w.r.t\onedot} \def\dof{d.o.f\onedot}
\def\etal{et al.\ }

\author{Andrej Janda$^1$, Pierre Merriaux$^2$, Pierre Olivier$^2$, and Jonathan Kelly$^1$ \\[3mm]
$^1$Space \& Terrestrial Autonomous Robotic Systems Laboratory, University of Toronto, Toronto, Canada \\
\texttt{\{andrej.janda,jonathan.kelly\}@robotics.utias.utoronto.ca} \\[1mm]
$^2$LeddarTech Inc., Qu\'{e}bec, Canada \\
\texttt{\{pierre.merriaux,pierre.olivier\}@leddartech.com}}

\maketitle
\begin{abstract}
    Advances in lidar technology have made the collection of 3D point clouds fast and easy.
    While most lidar sensors return per-point intensity (or reflectance) values along with range measurements, flash lidar sensors are able to provide information about the shape of the return pulse.
    The shape of the return waveform is affected by many factors, including the distance that the light pulse travels and the angle of incidence with a surface.
    Importantly, the shape of the return waveform also depends on the material properties of the reflecting surface.
    In this paper, we investigate whether the material type or class can be determined from the full-waveform response.
    First, as a proof of concept, we demonstrate that the extra information about material class, if known accurately, can improve performance on scene understanding tasks such as semantic segmentation.
    Next, we learn two different full-waveform material classifiers: a random forest classifier and a temporal convolutional neural network (TCN) classifier.
    We find that, in some cases, material types can be distinguished, and that the TCN generally performs better across a wider range of materials.
    However, factors such as angle of incidence, material colour, and material similarity may hinder overall performance.
\end{abstract}

\begin{figure}[t!]
\vspace{2mm}
\centering
\setlength{\fboxsep}{0pt}%
\setlength{\fboxrule}{1pt}%
\fbox{%
    \parbox[c]{\columnwidth - 2pt}{
        \includegraphics[width=\columnwidth - 2pt]{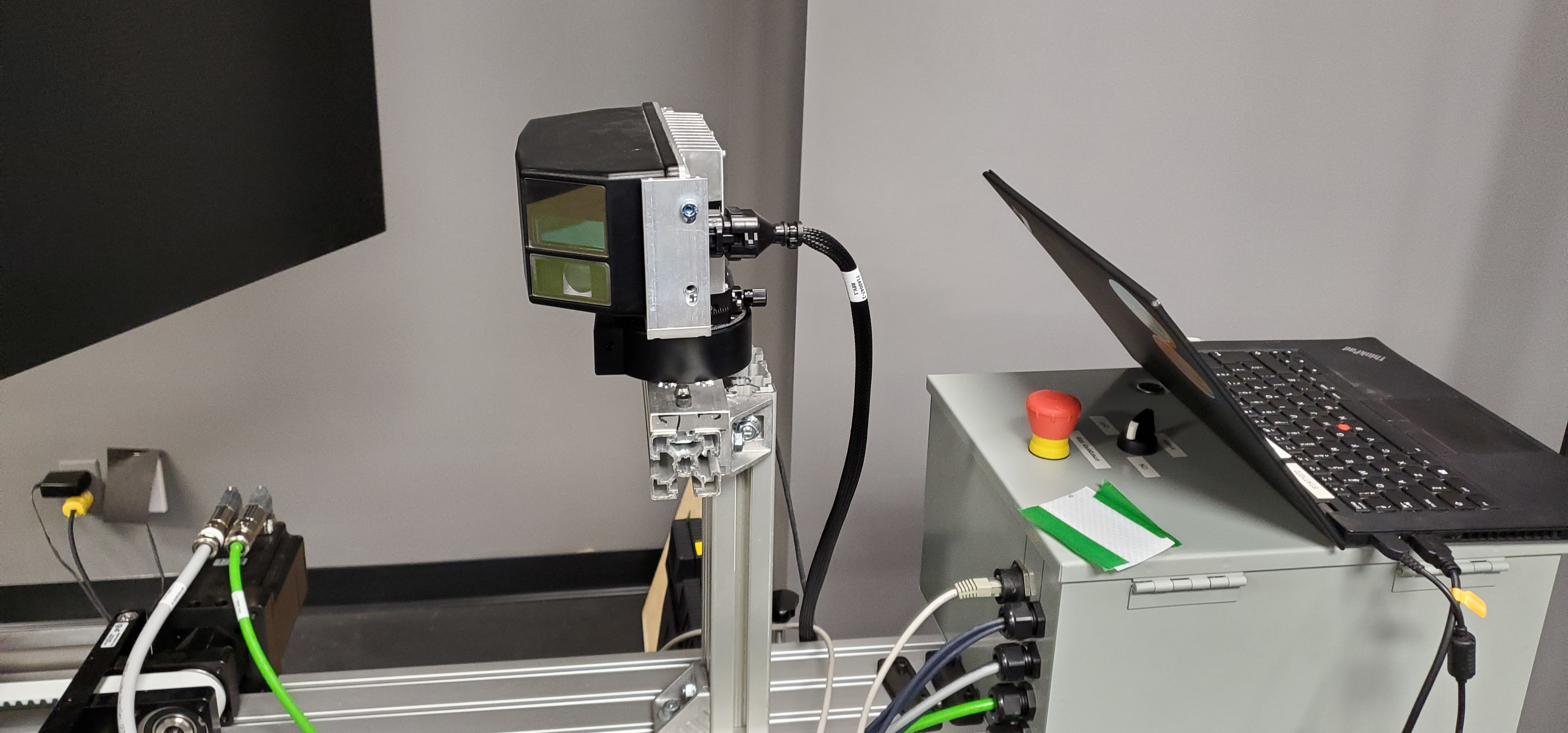}\\
        \includegraphics[width=\columnwidth - 2pt]{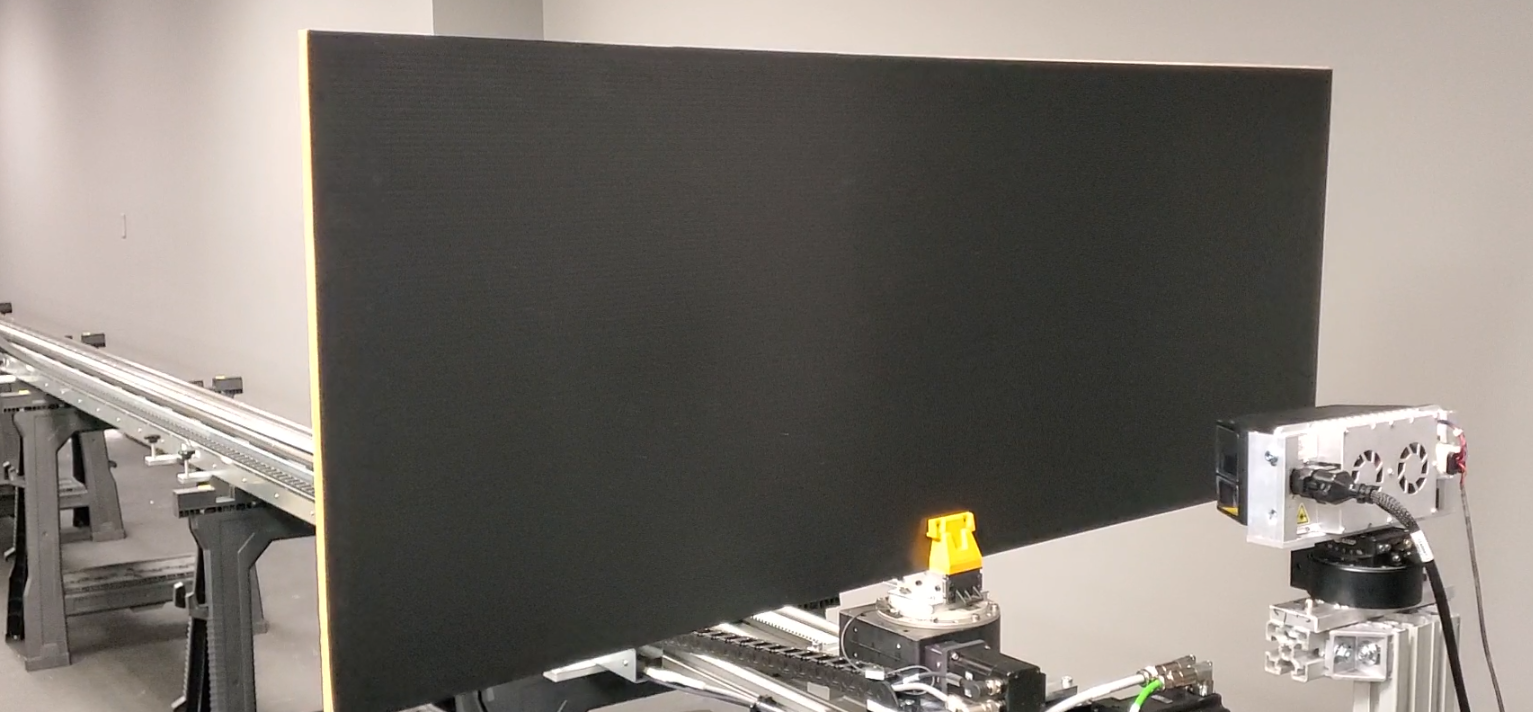}}}
\vspace{1mm}
\caption{Experimental setup for data collection. A full-waveform LeddarTech Pixell lidar sensor is affixed to a calibration bench, where a linear stage moves a rectangular board fitted with the material undergoing testing. The board can also be rotated about the vertical axis.}
\label{fig:experimental_setup}
\vspace{-4mm}
\end{figure}

\section{Introduction}

Point clouds form a key data modality because they can be used to accurately reconstruct the geometry of a 3D scene.
Reliable geometric reconstruction allows robots to interact meaningfully with the environment, carrying out path planning, obstacle avoidance, and object manipulation.
The value of point clouds, coupled with the increasing availability of 3D sensors (e.g., stereo cameras and lidar sensors), has made 3D data increasingly relevant to many robotics tasks.
Although there exist many ways to capture point clouds, lidar has proven to be a vital tool for many outdoor applications, where other sensors such as cameras have insufficient range and limited functionality under difficult lighting conditions.

Once a scene has been reconstructed, a common next step is to segment parts of the scene using semantically meaningful labels.
These semantic labels allow robots to interact with their environment under human direction.
Segmenting raw data into distinct semantic classes remains a challenging and open problem.
Recent work has shown that learning algorithms can be applied as an effective means to segment both image and point cloud data.
The difficulty of semantic segmentation motivates us to leverage any possible advantage to be gained from existing sensor technology.

\begin{figure*}
    \hfill
    \includegraphics[trim=0 0 0 0.5in,clip,width=7cm]{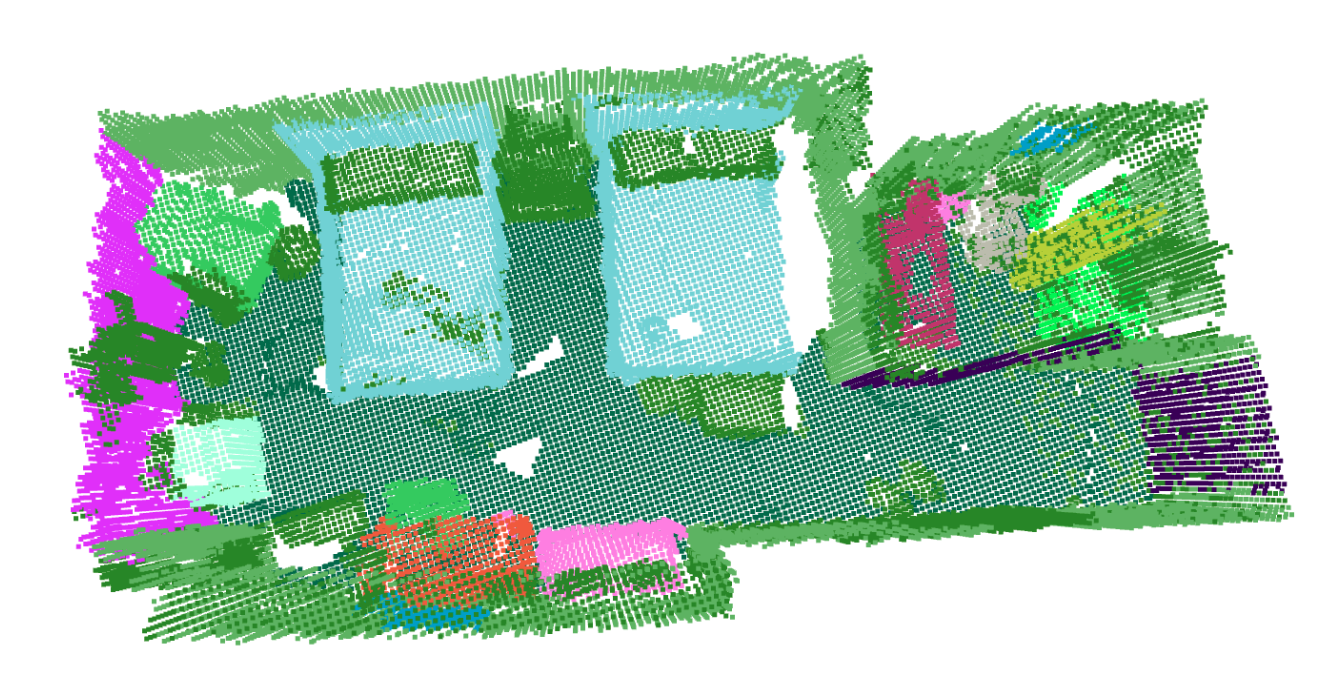}
    \hfill
    \includegraphics[trim=0.6in 0 2.4in 0,clip,width=5cm]{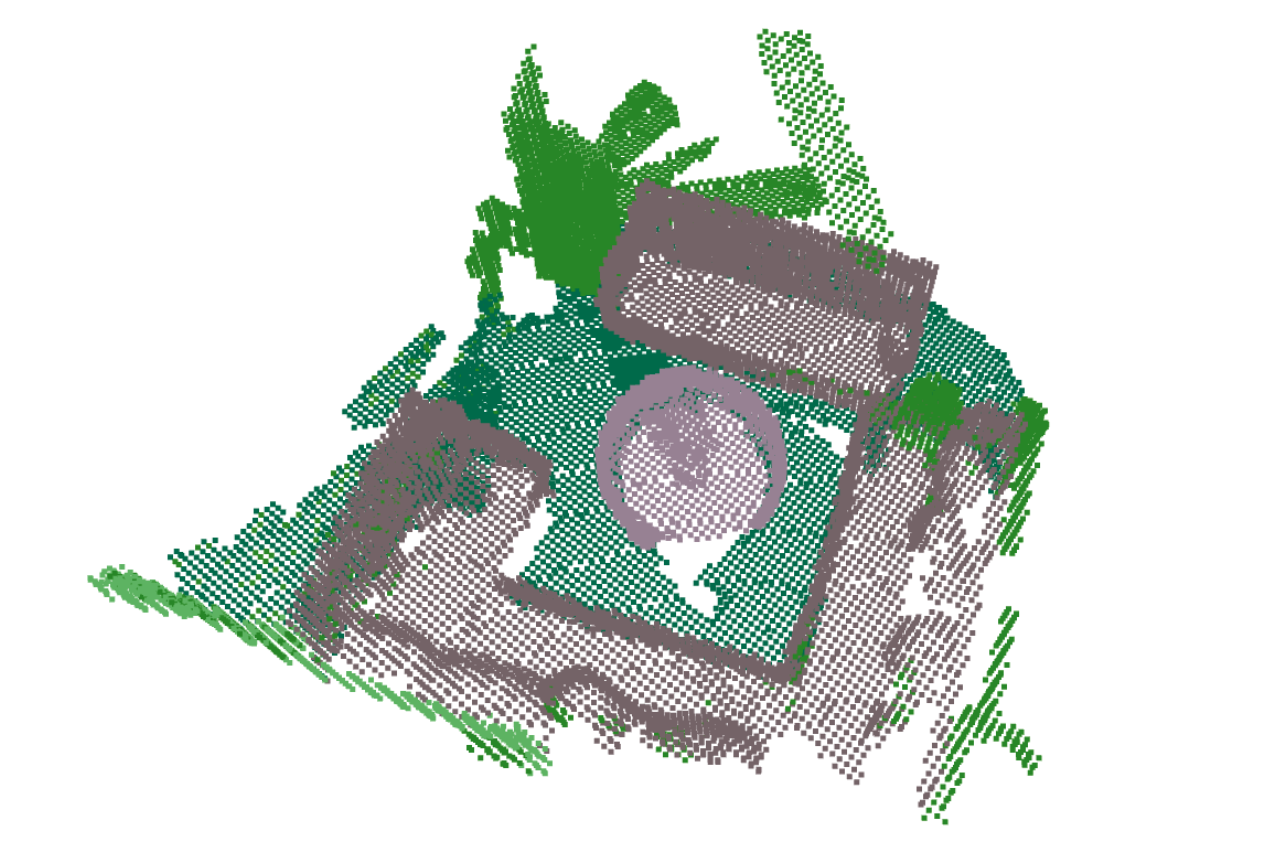}
    \hfill
    \includegraphics[trim=1in 0.5in 1.5in 0.5in,clip,width=5cm]{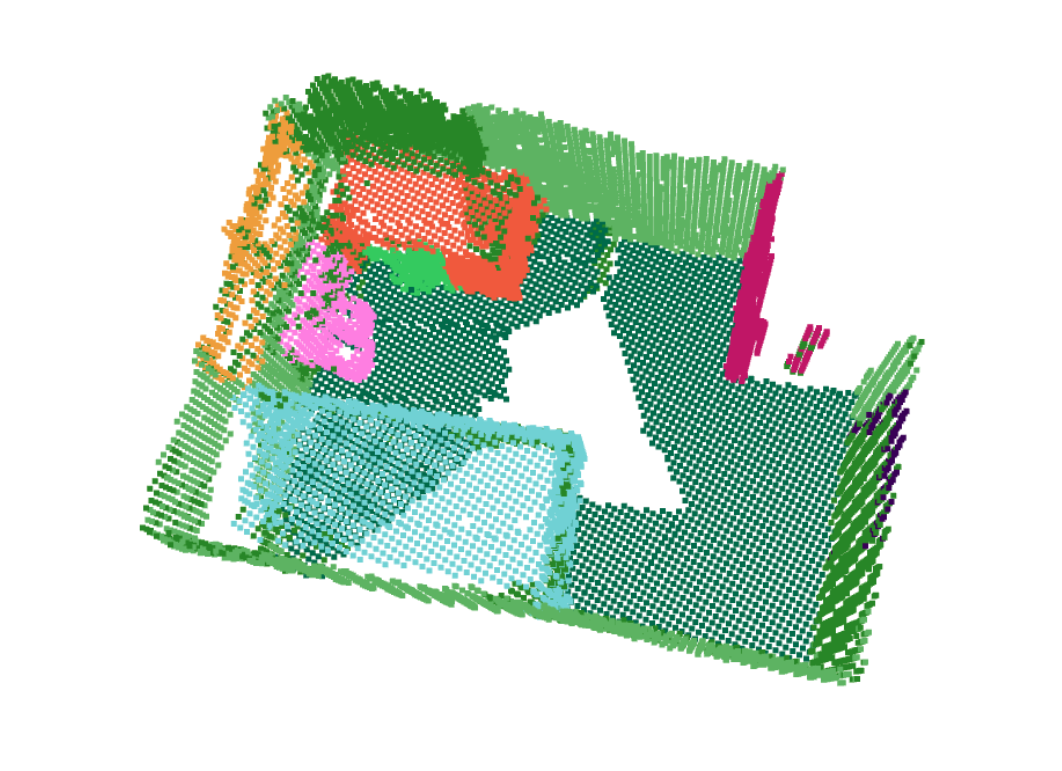}
    \hfill
    \caption{Visualization of material class labels added to reconstructed point clouds for three different scenes from the ScanNet dataset \cite{dai2017scannet}. Each colour represents a separate material class.}
    \label{fig:material_class}
    \vspace{-3mm}
\end{figure*}

One possible source of additional, useful information that has been mostly overlooked in the recent literature is the availability of full-waveform flash lidar data.
Reflection from an object in the scene induces a change in the lidar pulse waveform, which potentially contains information about the object itself.
Since the waveform is modified by interactions with the surface from which the signal is reflected, it may be possible to use the full waveform to determine surface material properties.
This application has yet to be fully explored and has the potential to improve the performance of existing semantic segmentation algorithms (by adding an additional feature channel).
We therefore investigate the viability of using full-waveform flash lidar data to determine material properties.
In short, this work makes the following contributions:
\begin{itemize}
    \item a detailed analysis of full-waveform lidar pulse data reflected from various materials;
    \item a demonstration of segmentation performance improvement with material class information;
    \item an evaluation of two learning-based models for material classification from lidar pulse data; and
    \item a breakdown of which parts of the waveform are most informative for material classification.
\end{itemize}

\section{Related Work}

The problem of extracting material class or properties from raw sensor measurements has been investigated for many common robotic sensing modalities.
Previous work on material classification from lidar data has focused on leveraging the reflected intensity of the laser pulse, measured by a single scalar value.
Song \etal \cite{song2012assessing} demonstrate that the intensity measurements can be used to distinguish various ground surface classes, such as asphalt and grass.
Yuan \etal \cite{yuan2020automatic} find that ensemble methods trained on material reflectance and roughness from laser data, along with colour information, can achieve a classification accuracy of various building materials above 97\%.
Similarly, Zahiri \etal \cite{zahiri2021characterizing} use multispectral images and laser beam intensity values to train an SVM classifier to distinguish between building foundation types.
Tatoglu and Pochiraju \cite{tatoglu2012point} apply first principles to compute the expected intensity value of each material using known reflectivity models.
The expected intensity is then compared against the observed intensity to compute a final material class prediction.
In this work, we instead focus on analyzing the full waveform of the returned lidar signal, which provides a much more rich source of information than a single intensity measurement.

Beyond lidar, other sensor modalities have also been used for material classification.
The methods described in \cite{weis2019material,weis2018oneshot} use full-waveform measurements from millimetre-wave radar sensors to classify material types with high accuracy.
Lu \etal \cite{lu2014feature} classify materials underground using a ground penetrating radar.
Saponario \etal \cite{saponario2015material} instead use the thermal properties captured from an infrared camera for material classification.
Some work has also been done to leverage tactile sensing for material classification \cite{baishya2016robust,jamali2010material}.
Zheng \etal \cite{zheng2016deep} use both haptic and visual information to extract material classes.

Although material classification has been studied extensively in the literature, the use of full-waveform data from flash lidar sensors has not been investigated extensively for this task.
Notably, waveform information has proven useful in other applications.
For example, the full lidar waveform has been used to: extract multiple detections from a single pulse when measuring tree canopies from topological scans \cite{blair1999laser}; improve the point density of lidar scans through super-resolution \cite{liu2019deep}; and to improve land-cover classification by fusing lidar scans with visual data \cite{wang_fusion_2015}.
Multiple works have also found that learning-based method are highly effective for airborne laser scanning (ALS) point cloud classification \cite{yang2017convolutional, ao2017one, zorzi2019full, chehata2009airborne}.
The proven utility of full-waveform measurements in other application is why we investigate, herein, how full-waveform data can be leveraged for scene understanding through material identification.

\section{Is Material Class Useful for\\ Semantic Segmentation?}
\label{sec:synthetic}

Before we examine the viability of distinguishing materials from full-waveform lidar data, we first investigate whether such information is useful for semantic segmentation tasks.
Since, to the best of the authors' knowledge, no dataset with both semantic and material labels exists, we choose to simulate this information by assuming a single material type for each semantic class.
We use ScanNet \cite{dai2017scannet} as our base dataset.
ScanNet is an indoor dataset of roughly 1,600 reconstructed scenes and contains semantic labels for 20 different classes.
Our assignment of material class to each semantic label is found in \Cref{tab:material_type_of_each_semantic_class}.
Our mapping tries to emulate the real-world material types of each class as closely as possible.
Multiple objects can therefore be mapped to the same material.
A visualization of the material classes for reconstructed point clouds from three different scenes is provided in \Cref{fig:material_class}.

For this experiment, we use the 3D sparse UNet architecture from \cite{choy20194d, choy2019minkowski} for semantic segmentation, with the same data augmentation and training parameters as in \cite{hou2021Exploring}.
At both training and inference time, the material class is appended to the existing colour information as an additional channel in the input feature map for each point.
We compare the impact that material labels have on semantic segmentation performance in \Cref{tab:synthetic_material_results}.
The results show a performance gain of $+$5.7 mIOU when material classes are added, compared to using colour information only.
This result motivates our work to extract the material class from full-waveform lidar data.

\begin{table}[t!]
\centering
\begin{threeparttable}
    \begin{tabular}{ c | p{0.3\textwidth}}
        \toprule
        \textbf{Material} & \textbf{Classes}                                              \\
        \midrule
        Drywall           & Wall                                                          \\
        Vinyl Laminate    & Floor                                                         \\
        Granite           & Counter                                                       \\
        Glass             & Window                                                        \\
        Paper             & Picture                                                       \\
        Enamel            & Bathtub, Toilet, Sink, Refrigerator                           \\
        Fabric            & Bed, Sofa, Curtain, Shower Curtain                            \\
        Wood              & Cabinet, Chair, Table, Door, Bookshelf, Desk, Other Furniture \\
        \bottomrule
    \end{tabular}
\end{threeparttable}
\vspace{2mm}
\caption{Simulated material association for each semantic class in the ScanNet dataset \cite{dai2017scannet}.}
\label{tab:material_type_of_each_semantic_class}
\vspace{-3mm}
\end{table}

\begin{table}[b!]
\vspace{-2mm}
\centering
\begin{threeparttable}
    \begin{tabular}{ c | c}
        \toprule
        \textbf{Available Features} & \textbf{mIOU} \\
        \midrule
        Colours                     & 62.5          \\
        Colours + Material Labels   & 68.2          \\
        \bottomrule
    \end{tabular}
\end{threeparttable}
\vspace{2mm}
\caption{Effect of adding synthetic material class labels to input features for semantic segmentation on the ScanNet dataset \cite{dai2017scannet}.}
\label{tab:synthetic_material_results}
\end{table}

\section{Experimental Setup}
\label{sec:experimental_setup}

In this section, we describe relevant details regarding our investigation.
We begin by discussing our data collection procedure, the materials we used, and our labelling methodology.
We then describe our approach to map the received lidar pulses to the correct material class.
Finally, we cover the specific evaluation metric applied to evaluate our results.

\begin{figure*}
    \centering
    \captionsetup[subfigure]{oneside,margin={-0.8cm,0cm}}
    \begin{subfigure}{8cm}
        \centering
        \hspace*{-1.2cm}
        \includegraphics[width=8cm]{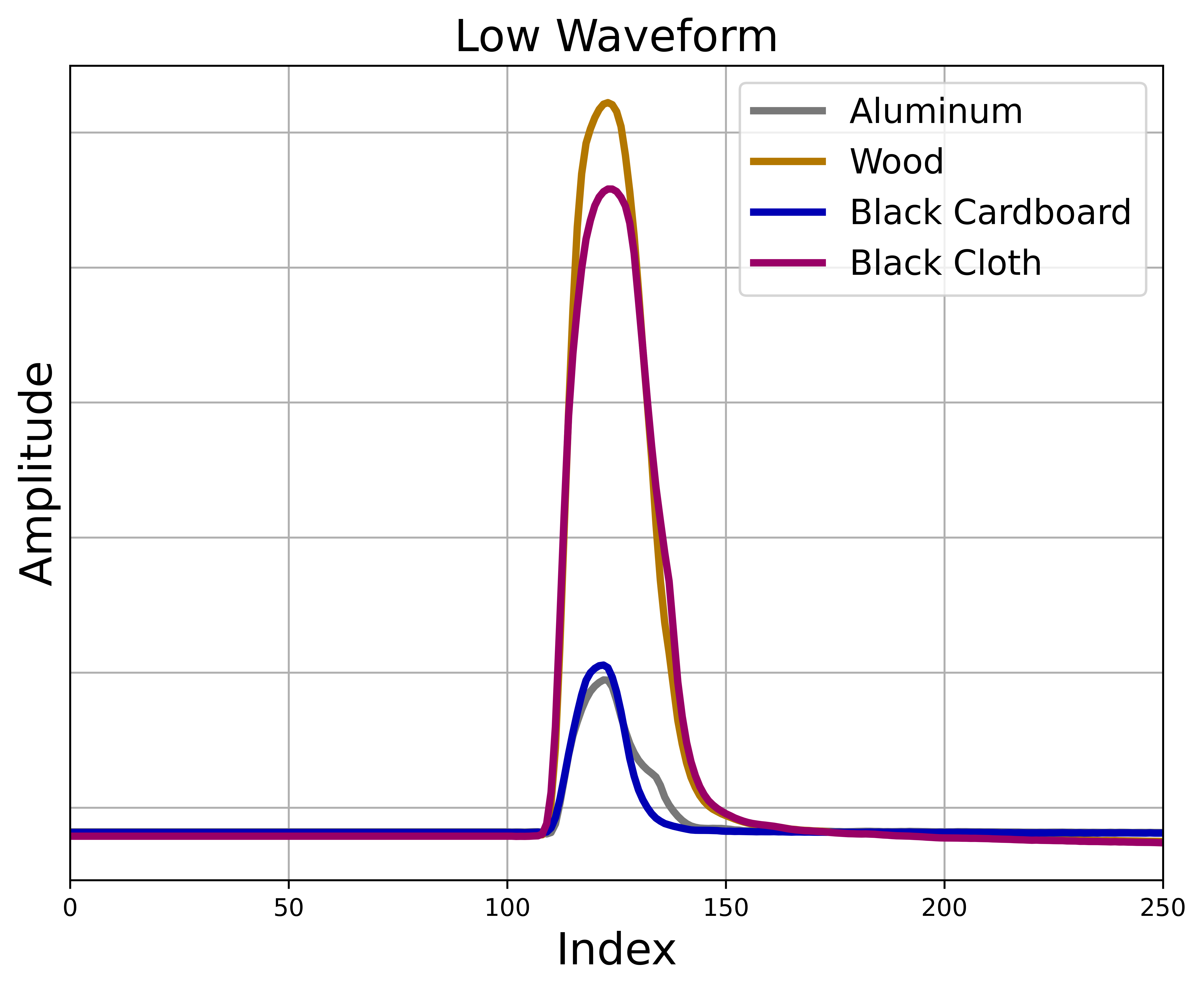}
        \caption{}
    \end{subfigure}
    \captionsetup[subfigure]{oneside,margin={1.1cm,0cm}}
    \begin{subfigure}{8cm}
        \centering
        \hspace*{0.2cm}
        \includegraphics[width=8cm]{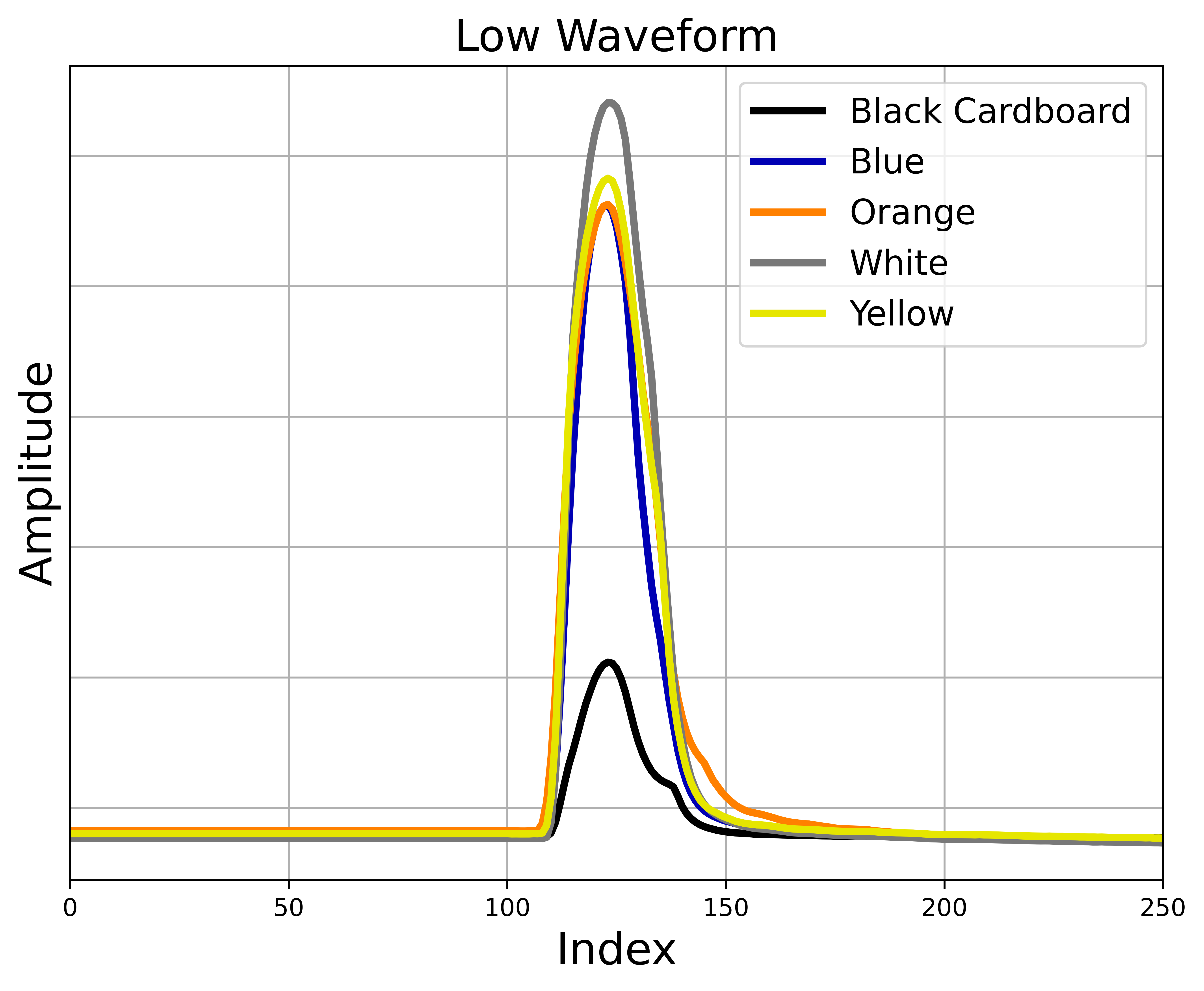}
        \caption{}
    \end{subfigure}
    \caption{Sample low-power waveforms averaged over multiple measurements using (left) samples from different materials and (right) samples from cardboard of different colours. Note that the amplitude is the raw proprietary value from the Pixell sensor and is therefore omitted for readability. The index axis of both graphs represents the position of consecutive measurements captured by the sensor, starting from the moment a lidar wave is first transmitted. The start of each graph contains a flat region, during which the transmitted wave has yet to return to the sensor after being reflected off of a surface.}
    \label{fig:waveform_samples}
\end{figure*}

\subsection{Data Collection and Labelling}
Measurements were collected on a test bench with a flash lidar unit in a fixed position. A test panel (board) was attached to a linear motion stage capable of moving forward (towards the lidar); the board could also be rotated about its vertical axis (see \Cref{fig:experimental_setup}).
For simplicity, the board was held at a fixed distance of 1 m and rotated in $15^\circ$ increments in the range $-60^\circ$ to $+60^\circ$.
For each rotation angle, five separate measurements were captured.
The lidar sensor uses both a high and low power signal.
We choose to focus on the low-power signal, since the close proximity of the board causes saturation of the sensor when using the high-power mode.
The low-power waveform is received as 256 individual measurements that we concatenate the input vector to our models.
Sample low-power waveforms for different materials are shown in \Cref{fig:waveform_samples}.
Measurements were taken with a single material affixed to the board, with the material covering the entire surface area of the board.
The materials tested were: aluminum, wood, black cardboard, and black cloth.
In addition to different materials, we also captured waveforms of reflections from cardboard of different colours.
The colours tested were: black, white, blue, orange, and yellow.
All sensor measurements that belonged to the immediate area around the board were labelled as `representing' the material type being measured.
Other points were treated as an unknown background class.

\begin{table}
    \centering
    \begin{threeparttable}
        \begin{tabular}{ c | l}
            \toprule
            \textbf{Parameter}  & \textbf{Value}                   \\
            \midrule
            Kernel size         & 1                                \\
            Dropout             & 0.05                             \\
            Channel Sizes       & 32, 32, 32, 64, 64, 64, 128, 128 \\
            Output layer        & Linear + SoftMax                 \\
            Batch Size          & 32                               \\
            Training Iterations & 4000                             \\
            Optimizer           & Adam                             \\
            Learning Rate       & 2e-3                             \\
            momentum            & 0.9                              \\
            \bottomrule
        \end{tabular}
    \end{threeparttable}
    \vspace{2mm}
    \caption{Network parameters of our TCN classifier model.}
    \label{tab:tcn_parameters}
    \vspace{-3mm}
\end{table}

\paragraph{Learning Models}
To map raw waveform data to a material class, we rely on learning-based models.
We use learning techniques because there is no simple decision criteria that can separate individual waveforms according to their respective material classes, as is evident from \Cref{fig:waveform_samples}.
Many of the waveform signatures for different classes overlap and vary extensively across different reflection angles.
Learning-based methods have proven to be adept at modelling complicated nonlinear decision boundaries.
Specifically, we choose models that are relatively small and fast to train, since, ultimately, the algorithms need to run in real-time to make their predictions accessible to downstream segmentation modules.
The first algorithm we investigate is the random forest (RF), which is an ensemble method that operates by finding consensus among multiple small decision trees.
For a more detailed background on random forests, we refer the reader to \cite{gerardRandom2016}.
Our model uses $100$ individual decision trees, each with a maximum depth of 50.
The second algorithm we investigate is the temporal convolutional neural network (TCN), which learns the parameters of successive 1D convolutional layers.
For a detailed background on TCNs, we refer the reader to \cite{BaiTCN2018}.
The parameters used for our model are listed in \Cref{tab:tcn_parameters}.
The output of last layer of the model is passed through a softmax classification function.
The model is trained using a cross entropy loss.

\paragraph{Evaluation Metric}
To evaluate the performance of each method for material classification, we use the \emph{Intersection Over Union (IOU)} metric, which is commonly employed to evaluate performance on semantic segmentation tasks.
The IOU measures both the precision (i.e., the proportion of predictions that were correct) and the recall (i.e., the proportion of data points that were correctly predicted) as a single value.
For problems spanning multiple classes (as is the case in this investigation), the metric is averaged across each class and is referred to as the \emph{mean IOU} or mIOU.

\begin{table*}[t!]
    \centering
    \begin{subtable}[t]{0.28\textwidth}
        \centering
        \begin{threeparttable}
            \begin{tabular}{ c | c c}
                \toprule
                \textbf{Model}       & \textbf{Angles} & \textbf{mIOU} \\
                \midrule

                \multirow{2}{*}{RF}  & 0               & 86.0          \\
                                     & All             & 65.0          \\
                \midrule
                \multirow{2}{*}{TCN} & 0               & 87.4          \\
                                     & All             & 64.1          \\
                \bottomrule
            \end{tabular}
        \end{threeparttable}
        \caption{Aluminum \& black cloth.}
        \label{tab:material_classification_1}
    \end{subtable}
    \hspace{1em}
    \begin{subtable}[t]{0.28\textwidth}
        \centering
        \begin{threeparttable}
            \begin{tabular}{ c | c c}
                \toprule
                \textbf{Model}       & \textbf{Angles} & \textbf{mIOU} \\
                \midrule

                \multirow{2}{*}{RF}  & 0               & 76.4          \\
                                     & All             & 44.3          \\
                \midrule
                \multirow{2}{*}{TCN} & 0               & 76.3          \\
                                     & All             & 49.7          \\
                \bottomrule
            \end{tabular}
        \end{threeparttable}
        \caption{Aluminum, black cloth, wood, and black cardboard.}
        \label{tab:material_classification_2}
    \end{subtable}
    \hspace{1em}
    \begin{subtable}[t]{0.28\textwidth}
        \centering
        \begin{threeparttable}
            \begin{tabular}{ c | c c}
                \toprule
                \textbf{Model}       & \textbf{Angles} & \textbf{mIOU} \\
                \midrule

                \multirow{2}{*}{RF}  & 0               & 57.1          \\
                                     & All             & 17.0          \\
                \midrule
                \multirow{2}{*}{TCN} & 0               & 49.8          \\
                                     & All             & 30.3          \\
                \bottomrule
            \end{tabular}
        \end{threeparttable}
        \caption{Different cardboard colours.}
        \label{tab:material_classification_3}
    \end{subtable}

    \caption{Material classification results.}
    \label{tab:temps}
\end{table*}

\begin{figure*}
    \begin{tabular}{ccc}
        \begin{subfigure}{6cm}
            \centering
            \includegraphics[width=6cm]{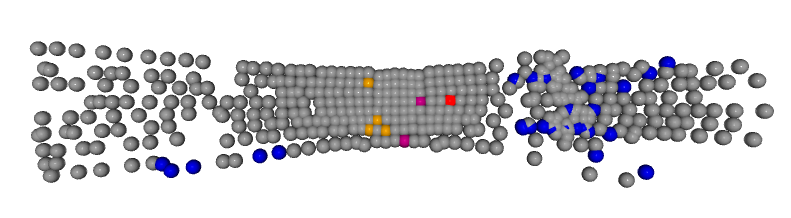}
            \vspace*{-0.5cm}
            \caption{Aluminum}
        \end{subfigure}
     & \begin{subfigure}{6cm}
           \centering
           \includegraphics[width=6cm]{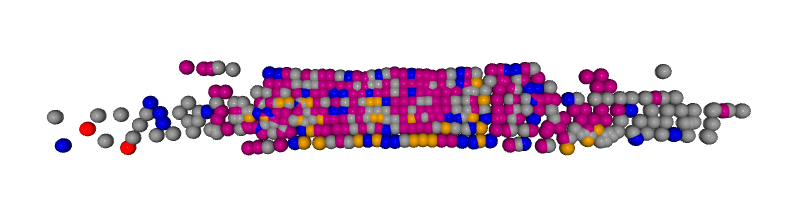}
           \vspace*{-0.5cm}
           \caption{Black Cardboard}
       \end{subfigure}
     & \multirow{2}{*}{
            \begin{subfigure}{4cm}
                \begin{tabular}{ll}
                    \textcolor{myGray}{$\blacksquare$}   & Aluminum        \\
                    \textcolor{myWood}{$\blacksquare$}   & Wood            \\
                    \textcolor{myBlue}{$\blacksquare$}   & Black Cardboard \\
                    \textcolor{myPurple}{$\blacksquare$} & Black Cloth     \\
                    \textcolor{red}{$\blacksquare$}      & Unkown
                \end{tabular}
            \end{subfigure}
        }                                                                                                                                                     \\
        \begin{subfigure}{6cm}
            \centering
            \includegraphics[width=6cm]{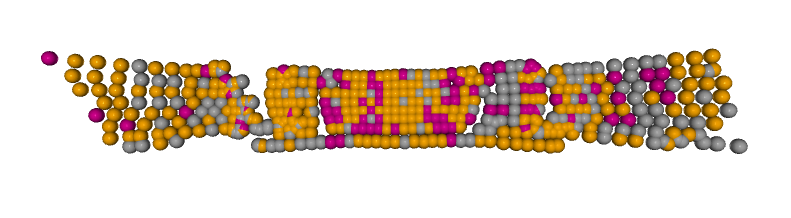}
            \vspace*{-0.5cm}
            \caption{Wood}
        \end{subfigure} & \begin{subfigure}{6cm}
                              \centering
                              \includegraphics[width=6cm]{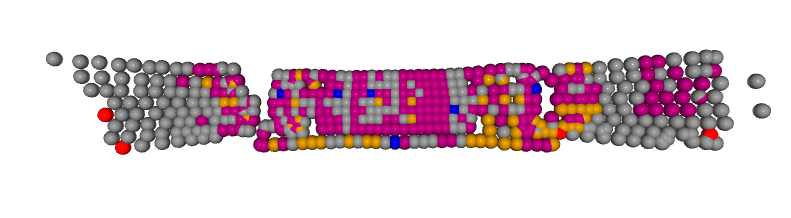}
                              \vspace*{-0.5cm}
                              \caption{Black Cloth}
                          \end{subfigure}                                                          \\
    \end{tabular}
    \caption{Visualization of segmentation results for the random forest model for all angles and materials. Each figure shows the segmentation of a board with a single material type.}
    \label{fig:visualization_all_materials}
    \vspace{-2mm}
\end{figure*}

\begin{figure}[b!]
\centering
\hspace*{-0.55cm}
\includegraphics*[trim=0 0 0 40pt,clip,width=9cm]{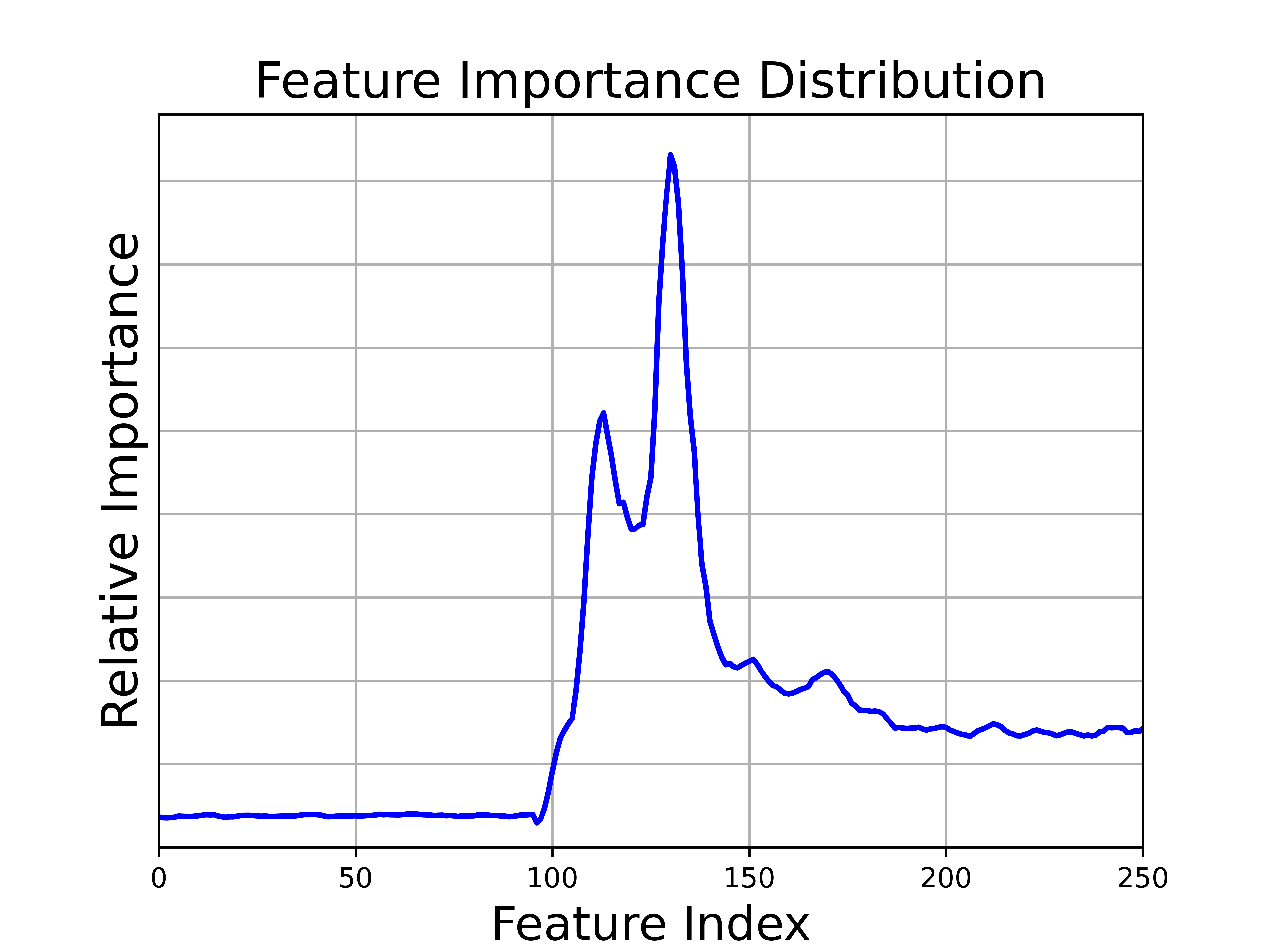}
\caption{Distribution of the relative importance of individual feature elements for material classification by a random forest.}
\label{fig:feature_importance}
\end{figure}

\section{Material Classification}
\label{sec:material_classification}

\begin{figure*}
    \begin{tabular}{ccc}
        \begin{subfigure}{6cm}
            \centering
            \includegraphics[width=6cm]{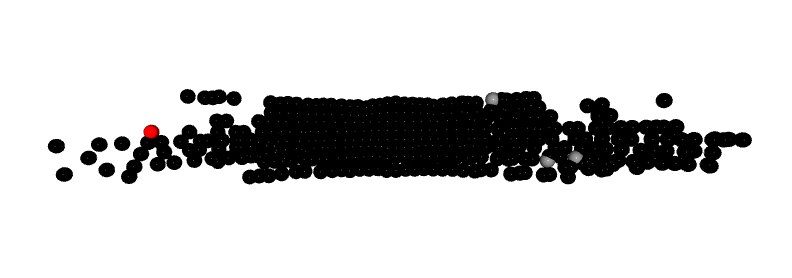}
            \vspace*{-0.5cm}
            \caption{Black}
        \end{subfigure}
         & \begin{subfigure}{6cm}
               \centering
               \includegraphics[width=6cm]{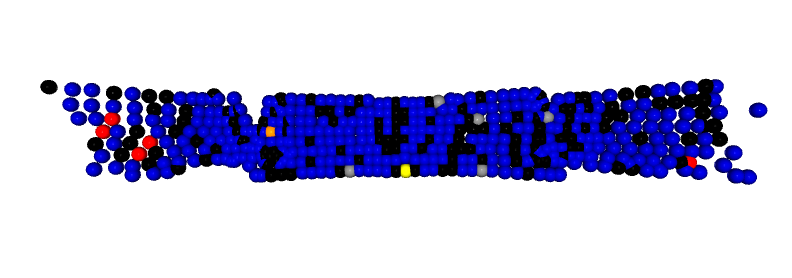}
               \vspace*{-0.5cm}
               \caption{Blue}
           \end{subfigure}
         & \multirow{2}{*}{
            \begin{subfigure}{4cm}
                \begin{tabular}{ll}
                    $\blacksquare$                       & Black  \\
                    \textcolor{myBlue}{$\blacksquare$}   & Blue   \\
                    \textcolor{myOrange}{$\blacksquare$} & Orange \\
                    \textcolor{myYellow}{$\blacksquare$} & Yellow \\
                    \textcolor{red}{$\blacksquare$}      & Unkown
                \end{tabular}
            \end{subfigure}
        }                                                                                                                                \\
        \begin{subfigure}{6cm}
            \centering
            \includegraphics[width=6cm]{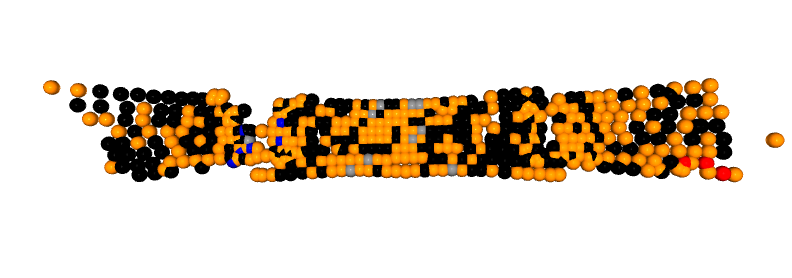}
            \vspace*{-0.5cm}
            \caption{Orange}
        \end{subfigure} & \begin{subfigure}{6cm}
                              \centering
                              \includegraphics[width=6cm]{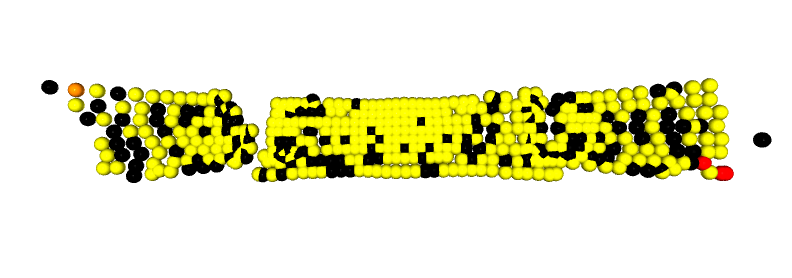}
                              \vspace*{-0.5cm}
                              \caption{Yellow}
                          \end{subfigure}                                                \\
    \end{tabular}
    \caption{Visualization of segmentation results for the random forest model for all angles and colours. Each figure shows the segmentation of a board with a single material type.}
    \label{fig:visualization_all_colours}
    \vspace{-2mm}
\end{figure*}

In this section, we analyze the performance of both RFs and TCNs when classifying material type from full-waveform data.
We split the material tests into three sets.
The first is a simple test using only aluminum and black cloth.
These materials were selected because they exhibit vastly different reflectivity and should be easy to distinguish (see \Cref{tab:material_classification_1}).
The second test utilizes all the materials that were available (see \Cref{tab:material_classification_2}).
Lastly, we examined whether the waveform could be used to distinguish between different colours of the same material (see \Cref{tab:material_classification_3}).
Each experiment was conducted using (1) no rotation (yaw), and (2) rotation from $-60^\circ$ to $+60^\circ$ in $15^\circ$ increments (indicated as \emph{All}).
Visualizations of the classified point cloud are shown in \Cref{fig:visualization_all_materials,fig:visualization_all_colours}.

Our results indicate that in the first and simplest case (\Cref{tab:material_classification_1}), the classification performance can be quite accurate, particularly when no rotation is considered.
The performance decreases when wood and black cardboard are added to the list of possible classes.
When all possible classes and rotations are considered, both models yield an mIOU bellow $50\%$, which would add little value to existing semantic segmentation methods.
Our results also show that both models struggle to distinguish between different colours of the same material, as seen in \Cref{tab:material_classification_3}.
Including all rotations of the different cardboard smaples resulted in the worst performance of all three experiments.
Therefore, we conclude that the waveforms from the same material type with different colours are essentially indistinguishable.
This is to be expected in part because the lidar return is affected by reflectivity and by colour.

In addition to the results above, we highlight a few key observations.
First, both the RF and the TCN perform similarly.
Since the TCN is a significantly more expressive model, we believe that the limitation lies within the data and not with the choice of a specific model.
Second, we find that the a non-zero yaw angle results in a dramatic decrease in model classification performance.
The angle of reflectance alters the return waveform, which may then have substantial overlap with the waveforms from other materials measured at different angles.

\section{Relative Feature Importance}
\label{sec:feature_importance}

A special property of random forests is that they can identify which parts of the feature vector are most influential during decision making.
The influence of each element is measured according to the frequency with which that element is chosen for `splitting' at each node of the individual decision trees.
We visualize this importance graphically in \Cref{fig:feature_importance}.
Compared to the waveforms in \Cref{fig:waveform_samples}, we see that the plateau at the start of the waveform has no impact on the final performance.
This is simply because no reflected signal has been received by the sensor, and so there is no useful information.
In contrast, the region after the peak has greater influence on the final result.
The `tail' of the waveform exhibits a varying signal before the amplitude reaches zero, and this region contains some limited information about the material.
The most important regions, however, are those where the waveform reaches its maximum amplitude.
We observe two peaks in the graph, one which relates to non-saturated amplitudes and another that captures saturation of the sensor.
The existence of saturation serves as a good differentiator between non-reflective materials (such as black cardboard) and more reflective materials (such as aluminum).
Further analyzing feature importance may improve feature selection, and could reduce the feature vector size as well as training and inference time.
We leave these steps as future work.

\section{Conclusion}

We set out in this short paper to determine whether distinguishing the material class of an object from full-waveform flash lidar measurements is possible.
We began by demonstrating the added value of material knowledge for semantic segmentation tasks.
We then described our real-world dataset and evaluated the performance of two different learning-based models for distinguishing material class from waveform data.
Our results indicate that, in some cases, for example when focussing on a few materials with very different reflectivity properties and at similar angles of incidence, material classification is a viable strategy.
However, as the number of distinct material classes and possible incidence angles grows, material classification becomes much more difficult and ultimately may be of limited use for segmentation tasks.

\bibliographystyle{IEEEtran}
\bibliography{venues_abbr,2023-janda-full-waveform-crv_short}

\end{document}